\documentclass[useAMS, referee, usenatbib]{biom}

\usepackage{natbib}
\usepackage{amsmath,amsfonts}
\usepackage{graphicx}
\usepackage{subcaption}
\usepackage{enumerate}
\usepackage{natbib}
\usepackage{url} 
\usepackage{booktabs}
\usepackage{tabularx}
\usepackage{lscape} 
\usepackage{bmpsize}
\usepackage{setspace}
\usepackage{mathrsfs}
\usepackage{amssymb}

\usepackage[hidelinks]{hyperref}
\hypersetup{
    colorlinks,
    linkcolor={red!50!black},
    citecolor={blue!50!black},
    urlcolor={blue!80!black}
}

\usepackage[utf8]{inputenc}
\usepackage{multirow,bigdelim, rotating}

\usepackage{zwcommands}
\usepackage{setspace}
\usepackage{mathtools}

\usepackage{dirtytalk}
\MakeRobust{\say}

\usepackage{xr}

\usepackage{hhline}
\usepackage{natbib}

\title[Neural Networks as Linear Regression: An Introduction for Statisticians]{Neural Networks as Linear Regression: An Introduction for Statisticians}

\author{Abigail Loe$^{1,*}$\email{aloe@macalester.edu}, 
Susan Murray$^{2}$, and Zhenke Wu$^{2}$\\
$^1$Department of Mathematics, Statistics and Computer Science, Macalester College,\\
Saint Paul, Minnesota, U.S.A.\\
$^2$Department of Biostatistics, University of Michigan, Ann Arbor, Michigan, U.S.A.}

\begin{document}





\pagerange{\pageref{firstpage}--\pageref{lastpage}} 
\volume{16}
\pubyear{2026}
\artmonth{October}




\label{firstpage}


\begin{abstract}
Neural networks are a commonly used prediction tool in computer science and statistics. However, the barrier to entry of this interesting field remains high, particularly for classical statisticians trained in a frequentist perspective. In this letter, we demystify neural networks by describing networks that approximate a linear regression and describe common customizations that provide a foundation for further study.
\end{abstract}



\maketitle


%

\section{Statistical Review of Neural Networks}
\label{s:nnet}
Neural networks are a popular way to model potentially complex functions with many variations and adaptations
\citep[see, for example][]{ LSTMlanguage, LSTMsleep, LSTMfuelcells}.
Although many features of neural networks have a clear statistical basis, jargon for describing these features are often quite different for computer science and statistical researchers, causing much unnecessary confusion. To familiarize statistical audiences with the neural network algorithm used in our research, we first introduce a neural network algorithm that corresponds to linear regression in Section \ref{s:nnet:linReg}. Then in Section \ref{s:nnet:lstmDesc} we describe a neural network algorithm appropriate for use  with the longitudinal data. Additional practical considerations of applying this neural network algorithm in our setting are given 
in Section \ref{s:nnet:modChoices}.

\subsection{Linear Regression as a Neural Network}\label{s:nnet:linReg}

In the language of computer scientists, neural networks are a user-specified architecture (model) with hidden layers of nodes, consisting of weights on the edges and biases (similar to intercept; different from the bias typically used in statistics) applied to model inputs for the purpose of predicting an outcome. Consider the regression model $g(E[\bY|\bZ_{1},\ldots,\bZ_{p}])=\mathbf{Z}^\top \beta$, where $\bY, \bZ_j, j = 1\ldots p$ are $n$-dimensional vectors, $\beta$ is of dimension $p\times 1$, and $\mathbf{Z}$ is the $(p+1)\times n$-dimensional design matrix, which includes the intercept term. Here and throughout this manuscript, we use bold face type to denote matrices or vectors. The $g(\cdot)$ function, or link function in statistical language, is typically the identity function for linear regression settings. Generally, we will use the term $g^{-1}(\mathbf{\eta})$ to denote the term that is used to minimize squared-error loss, i.e. $\{\mathbf{Y}-g^{-1}(\mathbf{\eta})\}^\top \{\mathbf{Y}-g^{-1}(\mathbf{\eta})\}$, where $\eta=\mathbf{Z}^\top \beta$ and $\eta$ is of dimension $n\times1$.

Translating neural network to statistical terminology: (1) model inputs are covariates, $\bZ_{j}$, $j=1,\ldots,p$; 
(2) weights are linear regression slope parameters $\beta_j, j=1,\ldots,p$; (3) bias corresponds to the linear regression model intercept, $\beta_0$; 
(4) a node is the  functional $g^{-1}(\mathbf{Z}^\top \beta)$, which reduces to $\mathbf{Z}^\top \beta$ in the setting mirroring linear regression with an identity link. Individual nodes make up the hidden layers of the neural network, termed hidden because neural network algorithms do not report estimates, $\hat{\mathbf{\beta}},$ just the vector of outcome predictions, $\hat{\bY}=g^{-1}(\mathbf{Z}^\top \mathbf{\hat{\beta}})$. Computer scientists refer to $g^{-1}(\cdot)$ as an activation function that estimates the desired outcome, often in several different layers within the algorithm.
Hereafter, we use the terms ``slope'' to refer to a weight, ``intercept'' to refer to bias, and ``parameter estimates'' to refer collectively to weights and biases used in the neural net contexts. 

Figure \ref{fig:basicnet} displays three increasingly more sophisticated neural networks in Panels (A), (B) and (C), where we include an additional subscript on $\eta$ to reflect changes across panels.
Panel (A) displays a linear regression model in a manner familiar to neural network researchers with a single hidden layer composed of one node; the corresponding statistical optimization function is displayed below the network. Covariates are represented by circles on the left of the diagram, while the linear predictor is displayed in the rectangular box. 

Researchers have found that an increase in the number of nodes in the hidden layer may improve prediction, particularly in the case where the relationship between covariates and outcome is more complex than a linear model. Figure \ref{fig:basicnet}, Panel (B) depicts a neural network with a single hidden layer composed of three nodes stacked on top of each other. The term, $\eta_B$, from Panel (B) is no longer a simple linear predictor as in panel (A), and involves parameter vectors $\alpha_1,\ldots,\alpha_3$ in addition to potentially different parameters, $\mathbf{\beta}_1,\ldots,\mathbf{\beta}_3$ applied to the same set of predictors, $\mathbf{Z}$. 
The form of $g_\ell^{-1}$, $\ell = 1, 2, 3$ may also vary across nodes. 

A network with two hidden layers is depicted in  Figure \ref{fig:basicnet}, Panel (C); hidden layers 1 and 2 have three and two nodes, respectively. The term, $\eta_C$, in Panel (C) has increased in complexity from Panel (B). Within $\eta_C$, readers can identify terms resembling $\eta_B$  fed into two different activation functions $h_1^{-1}(\cdot)$ and $h_2^{-1}(\cdot).$ Two issues emerge at this stage of neural network complexity that require comment.

First, the number of parameters involved in minimizing squared error loss based on the $\eta$ in panel C has more than doubled.  While the networks in Panels (B) and (C) have been found to potentially improve prediction compared to the network in Panel (A), parameter estimates are no longer guaranteed to be uniquely identifiable; it may be the case that there are multiple local minima of the loss function that are not the global minimum \citep{goodfellow2016deep}, as well as multiple different matrices of parameter estimates that may provide the same predictions. This is an open area of research, with mathematicians, statisticians and computer scientists attempting to find conditions that guarantee network architecture and parameter estimate uniqueness \citep[see, for example,][]{sussmann1992uniqueness, fefferman1994reconstructing, vlavcic2022neural}. In practice, however, most theorists recognize that it is generally sufficient if a local minimum is found; i.e. even local minima have sufficiently good predictive performance \citep{goodfellow2016deep}. 

The second issue that arises in Panel C is the potential for scaling differences to emerge between predicted outcomes generated within the hidden layers as the result of using so many different activation functions, $g_{\ell_1}^{-1}$ and $h_{\ell_2}^{-1}, \ell_1=1,2,3$ and $\ell_2=1,2$, within the neural network architecture. Computer scientists have approached this issue pragmatically by using the $\tanh$ function to rescale elements within the neural net architecture to be between a common -1 to 1 scale to reduce problematic differences. Some practitioners refer to the $\tanh$ operation as ``normalizing," though we should note that this is not normalizing in a statistical sense. 

Under certain assumptions, including non-colinearity of covariates, a linear regression squared error loss function has a closed form solution for parameter estimates. This is not typically the case for neural networks. Instead, iterative algorithms moving in smaller steps towards loss function (local) minima, via a hyperparameter known as the learning rate; each iteration is referred to as an epoch in neural network literature.
Popular algorithms are stochastic gradient descent, batch gradient descent, ``Adam" \citep{kingma2014adam}, and adaptive gradient descent \citep{duchi2011adagrad}, each with mini-batch variations that fall under the framework proposed by \citet{robbins1951minibatch}. For a comprehensive review, see \citet{bottou2018optimizationmethodslargescalemachine}.

Because of the large number of parameters built into neural network architectures, these algorithms are particularly susceptible to overfitting. 
A popular method used by both statisticians and computer scientists to prevent predictions being overfit to the data is to use different data cohorts for model training, validating, and testing. 
These terms have been used differently depending on the background of an individual researcher, so a quick summary of terms and techniques follows.
First, data are split into model building and model testing cohorts. The model building cohort determines the final architecture of the neural network using a $k$-fold cross-validation algorithm to evaluate architecture hyperparameters including, for instance, the number of hidden layers and the number of iterations used in the optimization algorithm. For a particular architecture defined using a set of hyperparameters, the $k$-fold cross-validation algorithm (1) splits the model building cohort into training and validation cohorts along a $(k-1):1$ ratio, (2) minimizes the loss function (squared-error loss for linear regression) in the training cohort to establish a working neural network, (3) uses the working neural network applied to the validation cohort to obtain predictions, $\hat{Y}$, and estimates of the loss function. Steps 1-3 are repeated for each of the $k$ data splits with loss function estimates across the $k$ validation cohorts averaged. This process is repeated for different sets of hyperparameters with a final architecture selection made based on smallest average estimated loss function across the validation cohorts. 

 Once the architecture hyperparameters are determined via $k$-fold cross-validation, the full model building cohort is used to rebuild the neural network (via re-minimizing the loss function in this cohort). The testing cohort is then applied to the final neural network to determine overall performance. For a detailed description of hyperparameter tuning and neural networks, see \citet{goodfellow2016deep}, Chapter 11, where decisions on optimization algorithm learning rate and number of epochs are discussed at length.
We will later summarize how these three groups may be more flexibly defined in the context of longitudinal outcomes in Section \ref{s:nnet:modChoices}.

To illustrate a typical process, we now describe the process of fitting the linear regression neural network depicted in Figure \ref{fig:basicnet}, Panel (A) on training data, though the process generalizes to the architectures displayed in Panels (B) and (C). For each epoch, $a = 1,2, \ldots n_a$ (where $n_a$ is often between 10 and 200):
\begin{enumerate}
    \item Calculate fitted values of $\hat{\mathbf{Y}}^{(a-1)}$ for the linear regression loss function $\hat{\mathbf{Y}}^{(a-1)} =\bZ^\top \Tilde{\mathbf{\beta}}^{(a-1)}$, where $(a)$ tracks the epoch. For the first epoch, traditionally $\Tilde{\mathbf{\beta}}^{(0)}=0$, though some authors note that sufficient convergence to a local minima can depend on the initial starting values \citep{wang1993stoppingrules}.
    \item Calculate the updated loss function values. For linear regression, this would be squared error loss with the formula $\mathcal{L}(\mathbf{Y},\hat{\mathbf{Y}}) = (\mathbf{Y}-\hat{\mathbf{Y}}^{(a-1)})^\top(\mathbf{Y}-\hat{\mathbf{Y}}^{(a-1)})  = \{\mathbf{Y} - g^{-1} (\bZ^\top \Tilde{\mathbf{\beta}}^{(a-1)} \}^\top\{\mathbf{Y} - g^{-1} (\bZ^\top \Tilde{\mathbf{\beta}}^{(a-1)} \}$, where $g^{-1}$ is typically taken to be the identity function. More generally, loss functions mimic those seen in generalized linear models, reflecting the assumed distribution of $\mathbf{Y}$.
    \item 
    Take partial derivatives of the loss function with respect to each parameter $\beta_0, \ldots, \beta_p$ to determine the gradient of the loss function. Update $\Tilde{\mathbf{\beta}}^{(a-1)}$ to $\Tilde{\mathbf{\beta}}^{(a)}$ in a step size corresponding to the learning rate and optimization algorithm of your choice. For example, with stochastic  gradient descent, $\Tilde{\mathbf{\beta}}^{(a-1)} =\Tilde{\mathbf{\beta}}^{(a)} - \triangledown \frac{\partial}{\partial \Tilde{\mathbf{\beta}}^{(a)}} \mathcal{L}$, where $\triangledown$ is the learning rate, and $\frac{\partial}{\partial \Tilde{\mathbf{\beta}}^{(a)}} \mathcal{L}$ is the partial derivatives of the loss function with respect to each parameter.
    \item Repeat steps 1-3 until $n_a$ is reached.
\end{enumerate}



\subsection{Useful Extensions to Bare-bones  Networks}\label{s:nnet:modChoices}
Neural networks have many proposed extensions. In this section, we selectively describe (1) common parameters available from the off-the-shelf Python module \texttt{torch.nn}, and (2) a few additional neural network design choices in common use for tailoring analyses.  
Options in \texttt{torch.nn} include: $\eta$ (computer scientists call this hidden state), parameter pruning (computer scientists call this the drop-out proportion), number of network layers, and the encoding of similarities within individuals via embeddings. 


A hyperparameter favored by computer scientists is allowing $\eta_{t}$ to be extended from a scalar to a matrix in $\mathbb{R}^{n(t)\times h}$, with columns $\eta_t^{(1)}, \eta_t^{(2)}, \ldots \eta_t^{(h)}$, and similarly $\bZ_i(t)$ to be extended from a vector in $p\times 1$ to a matrix of dimension $\mathbb{R}^{p \times n(t)}$. 
Computer scientists call $h$ the hidden dimension. For dimensions $h>1$, the inverse link is applied to a linear combination of the $h$ columns of $\eta_t$, i.e. $g^{-1}(\gamma_0+ \gamma_1 \mathbf{\eta}_t^{(1)}+ \ldots \mathbf{\eta}_t^{(h)})$. 

\begin{figure}
    \centering
    \includegraphics[width =\textwidth]{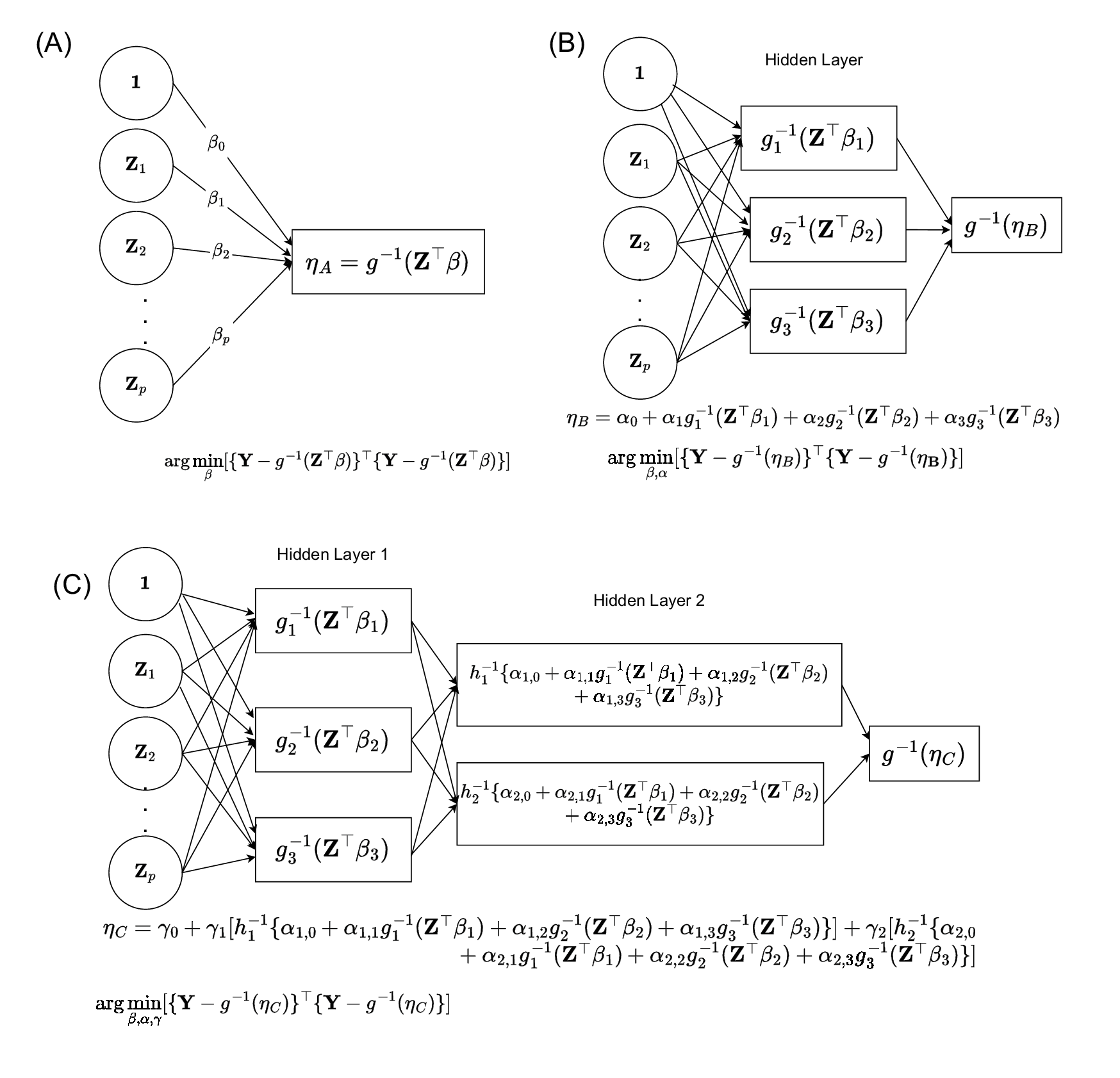}
    \caption{Three fully connected neural network architectures of increasing complexity, with squared error loss. Vectors of covariates are depicted in circles on the left, while nodes containing estimable parameters are displayed in rectangular boxes. Networks are called ``fully connected" because each covariate is passed to each node in the first layer, and for networks with more than one layer, outputs from previous layers are passed to all subsequent layers. Panel (A) displays an architecture with one hidden layer containing one node, akin to linear regression in the special case where $ g$ is the identity link. Panel (B) displays a network architecture with three nodes in one hidden layer, that are ultimately combined to obtain the linear component, $\eta_B$. As Panel (C) contains multiple hidden layers, Panel (C) proposes a network that may also be termed a ``deep learner."\\
    Alt. text: The figure containing three different neural networks. Panel (A) contains the simplest, while Panels (B) and (C) increase in complexity. Under each network architecture, the corresponding loss function is displayed.}
    \label{fig:basicnet}
\end{figure}

Beyond increasing $h$, computer scientists have also found that vertically stacking additional architecture at time $t$  increases prediction performance \citep{graves2014generatingsequencesrecurrentneural}. General recommendations are for at most 4 layers of architecture to be ``stacked" on top of each other \citep[as used in][]{sutskever2014sequencesequencelearningneural}. For instance, the entire architecture at time $t$ seen in Figure \ref{fig:basicnet}C might be stacked on top of a similar architecture, where the $\eta_C$ elements taken from the right side of the top stacked architecture play the role of the covariate inputs ($Z_{p}$ terms) on the left of the bottom stacked architecture. Additional subscripts on parameters and link functions in the bottom stacked architecture would complete this version of stacking architecture. 

The final two off-the-shelf options include (1) regularization via drop-out and (2) embeddings. Regularization via drop-out refers to  randomly forcing a proportion of parameters, $p_d$, to 0 during each epoch of training \citep{srivastava2014dropout, Zaremba2014EmbeddingRegularizer}. Subject (and potentially time point) embeddings are a concept borrowed from text mining \citep{sutskever2014sequencesequencelearningneural, Mikolov2012embeddings, mikolov2013word2vec}. For each individual, a dimension $b$ vector of embedding values may be allocated to quantify similarity of outcomes experienced by individuals. These embedding values are treated as parameters to be estimated in addition to other neural network outputs, and are available after training is complete.

Training, validation and testing cohorts described in Section \ref{s:nnet:linReg} may be used in a grid-search to optimize the number of training iterations selected from the set, $\{1,\ldots,n_a\},$ and the learning rate, $\triangledown$ selected from a size $m$ grid of candidate learning rates, $\{\triangledown_1, \triangledown_2\ldots, \triangledown_m\}, $.
All hyperparameters may be selected via a grid search algorithm, though a large number of points in a grid increases significantly increases computation time. As such, we chose to focus on learning rate and number of training epochs, two hyperparameters generally recognized to have an outsized effect on network reliability \citep{goodfellow2016deep}

Loss function decreases for each additional training iteration are expected in the training set, but not necessarily in the validation set. When increasing the number of training iterations from 1 to $n_a$ in the validation set for a particular learning rate, $\triangledown_\ell$, $\ell=1,\ldots,m$ if $\max_p$ consecutive increases in the loss function are seen, $n_{opt}$ is set to the number of training iterations, prior to the start of when these increases were seen (called early stopping in computer science literature); otherwise $n_{opt}=n_a.$ The value of $\max_p$ is a user chosen value traditionally between $5$ and $10$ that is called the tolerance or patience in computer science literature. Further details on early stopping may be found in \citet{prechelt1998automatic}. Validation loss and recommended number of epochs are recorded for each candidate learning rate, with final learning rate and recommended epochs, $\triangledown_{opt}$ and $n_{opt}$ respectively, selected based on the pairing that yields minimal loss in the validation set. 

\backmatter


\section*{Acknowledgments}
\textit{Conflict of Interest:} None declared.

\bibliographystyle{biom} 
\bibliography{mybib}

\appendix




\label{lastpage}

\end{document}